\title{Pose Estimation Based on 3D Models}
\author{Chuiwen Ma, Hao Su, Liang Shi}
\date{}
\documentclass[twocolumn]{article}
\usepackage{graphicx}
\usepackage{color}
\usepackage[top=1.1in, bottom=1.1in, left=1in, right=1in]{geometry}
\makeatletter
\def\@maketitle{%
	\newpage
	\vspace*{-\topskip}      
	\begingroup\centering    
	\let \footnote \thanks
	\hrule height \z@        
	{\LARGE \@title \par}%
	\vskip 1.5em 
	{\large
		\lineskip .5em 
		\begin{tabular}[t]{c}%
			\@author
		\end{tabular}\par}%
	\vskip 1em 
	{\large \@date}%
	\par\endgroup            
	\vskip 1.5em             
}
\makeatother
\begin{document}
\maketitle

\section{Introduction}\label{intro}
This project aims to estimate the pose of an object in the image. Pose estimation problem is known to be an open problem and also a crucial problem in computer vision field. Many real-world tasks depend heavily on or can be improved by a good pose estimation. For example, by knowing the exact pose of an object, robots will know where to sit on, how to grasp, or avoid collision when walking around. Besides, pose estimation is also applicable to automatic driving. With good pose estimation of cars, automatic driving system will know how to manipulate itself accordingly. Moreover, pose estimation can also benefit image searching, 3D reconstruction and has a large potential impact on many other fields. \\

Previously, most pose estimation works were implemented on manually captured and labeled dataset, using multi-camera or depth camera. However, to create such a dataset is extremely time-consuming, laborsome, and also error-prone. Therefore, limited information can be learned from those datasets, and research-specific datasets lead to poor comparability among previous works. In this project, we instead utilized the power of 3D shape models. To be specific, we built a large, balanced and precisely labeled training dataset from ShapeNet [\ref{bib:shapenet}], a large 3D model pool which contains millions of 3D shape models in thousands of object categories. By rendering 3D models into 2D images from different viewpoints, we can easily control the size, the pose distribution, and the precision of the dataset. A learning model trained on this dataset will help us better solve the pose estimation task.\\

In this work, we proposed a pose estimation system based on rendered image training set, which predicts the pose of objects in real image, with knowledge of object category and tight bounding box. Although the approach is generic, we chose chair to be our primary research object. Our system takes a properly cropped chair image as input, and outputs a probability vector on discretized pose space. Given a test image, we first divide it into a $N\times N$ overlapped patch grid. For each patch, a multi-class classifier is trained to estimate the probability of this patch to be pose $v$. Then, scores from all patches are combined to generate a probability vector for the whole image.\\

Although we created a larger and more precise training dataset from rendered images, there is an obvious drawback of this approach --- the statistical property of the training set and the test set are different. For instance, in the real world, there exists a prior probability distribution of poses, which might be non-uniform. Furthermore, even for feature complexity, real image features might be more diverse than rendered image features. In this project, we also focused on information transmission between 2D images and 3D models, therefore proposed a method to iteratively learn from classification results and in return improve classification algorithm. This novel approach revised the influence of different prior probability distribution in training and test set. Details and experiment results are shown in the following sections.

	\subsection{Related Works}
	Object pose estimation is a classical problem in computer vision. In general, there are two typical research lines: one based on 2D representation and the other based on 3D information.\\
	
	Among 2D based researches, [\ref{bib:lepetit},\ref{bib:haralick},\ref{bib:gold}] rely on point matching, which is now outdated. By linking together diagnostic parts of object from different views, [\ref{bib:silvio}] represents an object category as a collection of view-invariant regions. Sun \textit{et al.} [\ref{bib:sunsu}] and Su \textit{et al.} [\ref{bib:suhao}] used a generative approach to group local features into parts and then learn part locations across viewpoints. [\ref{bib:ozuysal}] used a SIFT-like [\ref{bib:SIFT}] spatial pyramids of histograms feature to train a SVM classifier for each discrete pose. Inspired by Deformable Part Model's [\ref{bib:DPM}] success, [\ref{bib:lopez}] trained a DPM  using a semi-latent approach, where the components correspond to discrete viewpoints. [\ref{bib:is2d}] used convolutional neural network features for the task of pose estimation. [\ref{bib:redondo}] proposed a Hough Forest based method for simultaneous object detection and continuous pose estimation. Those widely different works above, although gained some achievements, are not learning from structural information of objects like human.\\
	
	Recently, 3D model based approach achieved good performance on pose estimation task. [\ref{bib:pepikPDM}] extended deformable part models to 3D, where part appearances and spatial deformations are represented in 3D. Using an readymade approach, [\ref{bib:zia}] first obtained a rough localization and viewpoint of the object, and then estimated a continuous pose by using annotated 3D CAD models. Hejrati \textit{et al.} [\ref{bib:hejrati}] estimated poses of cars using an explicit 3D shape model and viewpoint which is learned from structure-from-motion (SFM). In general, methods above rely on sophisticated handling of 3D models, due to the limitation of model amount.\\
	
	Up to our knowledge, there is no previous work that utilizes large 3D model database to solve pose estimation task. 

\section{Data Collection and \\Processing}\label{data}
	\subsection{Training Data}\label{trainingdata}
\begin{figure}[h]
		\centering
		\includegraphics[width=0.24\textwidth]{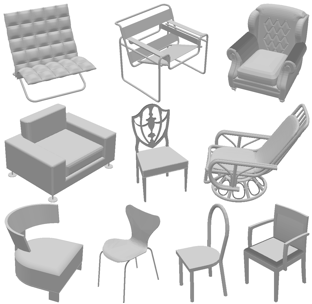} 
		\includegraphics[width=0.24\textwidth]{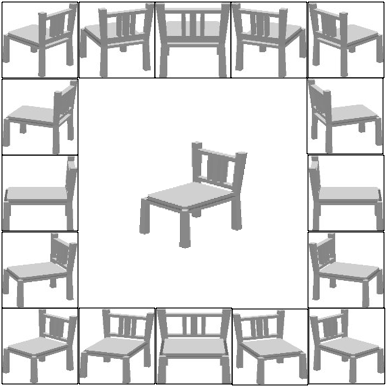}
		\caption{Chair models and rendering process}
		\label{fig:shapenet}
	\end{figure}

	As we mentioned in Section \ref{intro}, we collected our training data from ShapeNet, an emerging 3D shape model database. With 9,135135 semantically annotated 3D models in thousands of object categories, ShapeNet could provide abundant information for many vision tasks. In our task, we utilized those 5057 chair models in ShapeNet. For each model, we rendered it on 16 viewpoints, evenly distributed on the horizontal circle, shown in Figure \ref{fig:shapenet}.
	
	We chose 4000 models, accordingly 64,000 images to build the training dataset, and leave the rest 1057 models to be our rendered image test set (validation set). Before extracting image features, we first resize the images to $112 \times 112$ pixels, and then divide it into $6 \times 6$ overlapped patch grid, with patch size $32 \times 32$ and patch stride 16 on both axes. After preprocessing, we extract a 576 dimensional HoG feature ${[\ref{bib:hog}]}$ from each patch, so the whole image can be represented by a 20736 dimensional feature vector. Those 64,000 feature vectors constituted our training dataset.
	\begin{figure}[h]
		\centering
		\includegraphics[width=0.22\textwidth]{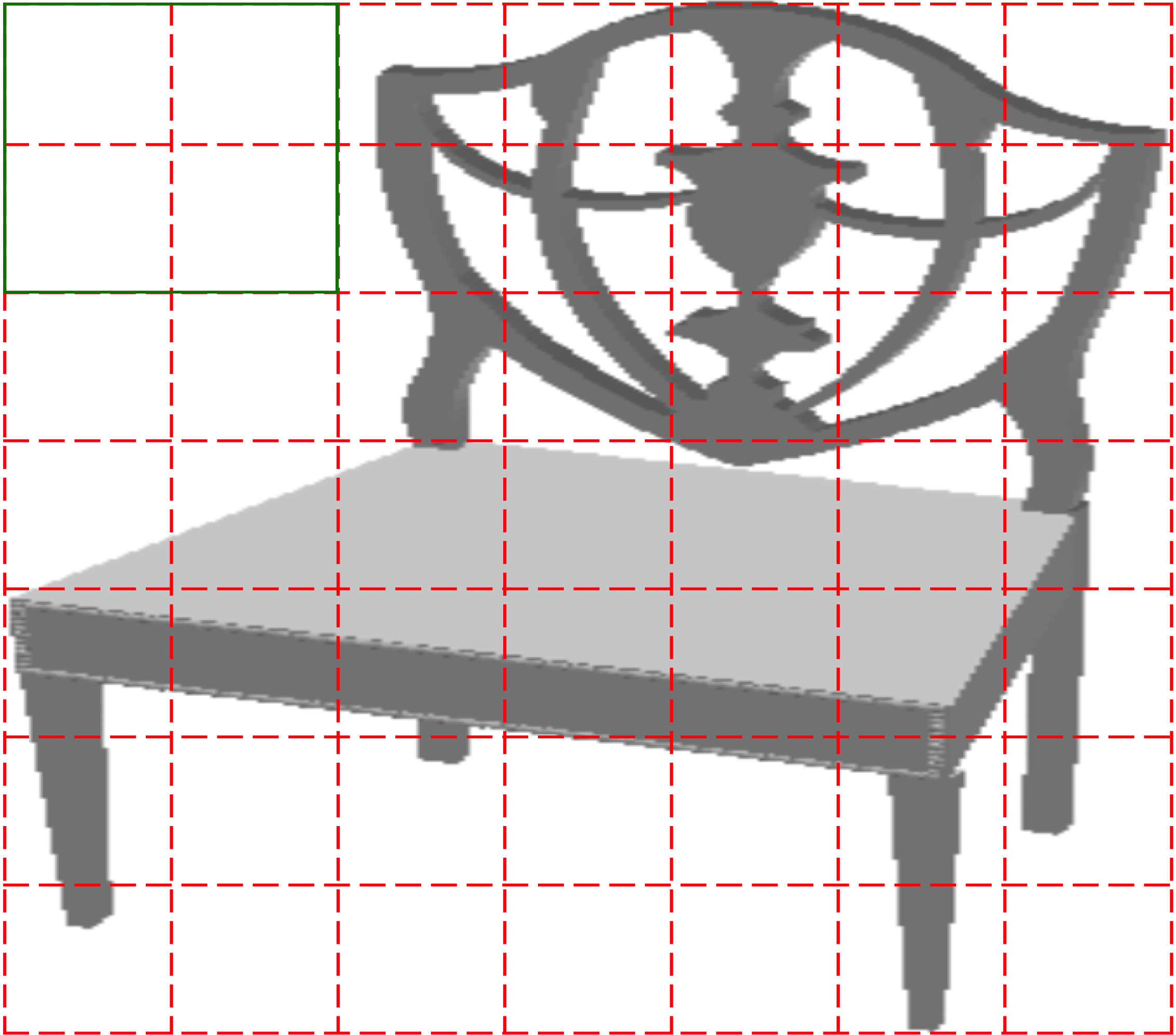} 
		\includegraphics[width=0.22\textwidth]{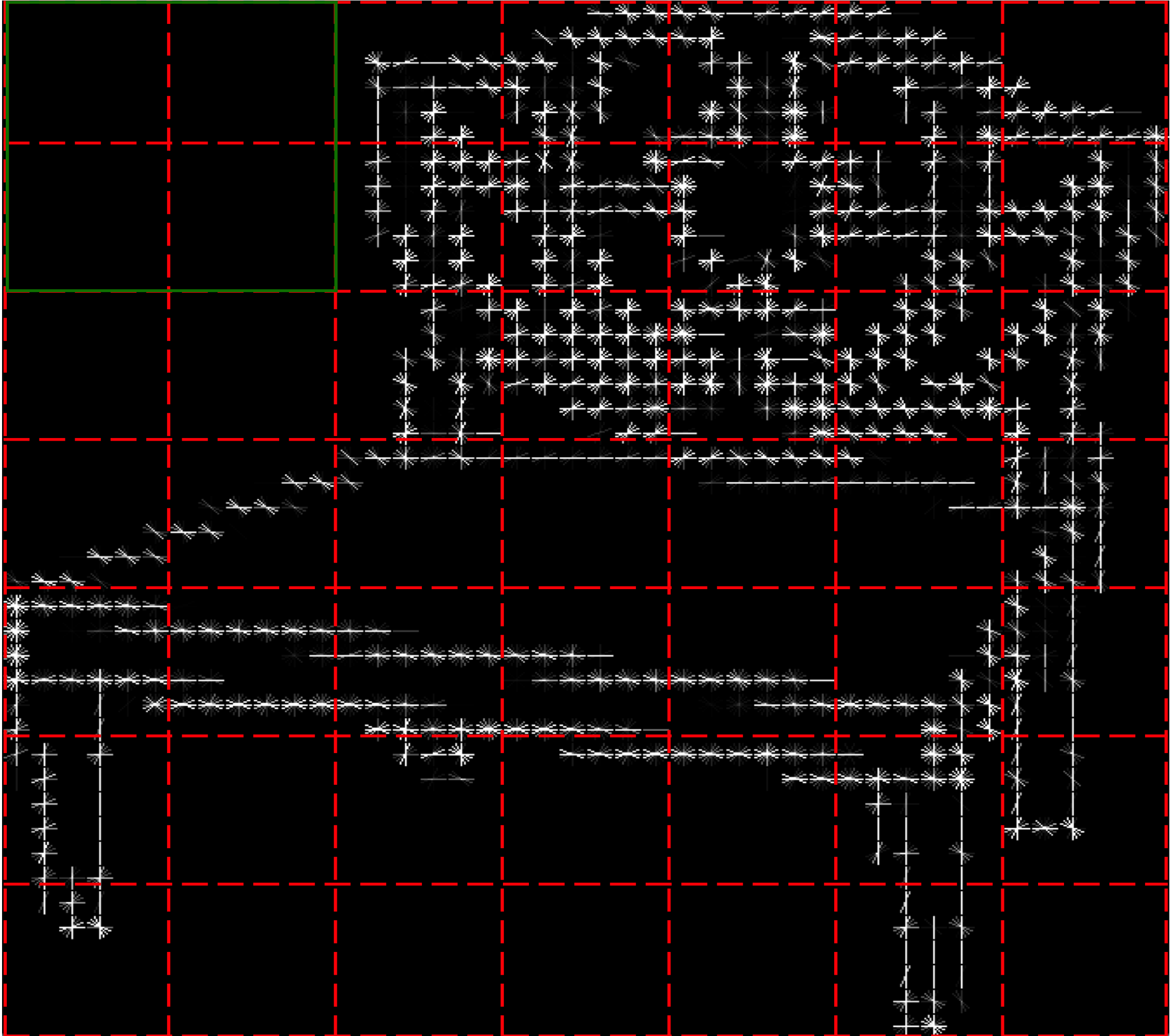}
		\caption{Image preprocessing and feature extraction}
		\label{fig:preprocessing}
	\end{figure}
	
	\subsection{Test Data}
	To comprehensively evaluate the performance of our learning algorithm, we built three different test sets with increasing level of test difficulty. They are rendered image test set, clean background real image test set and cluttered background real image test set.\\
	
\begin{figure}[h]
		\centering
		\includegraphics[width=0.22\textwidth]{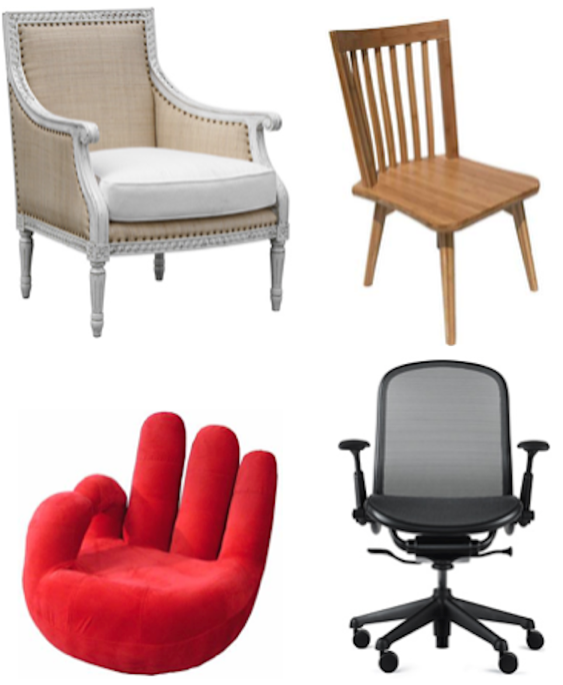}
		\includegraphics[width=0.26\textwidth]{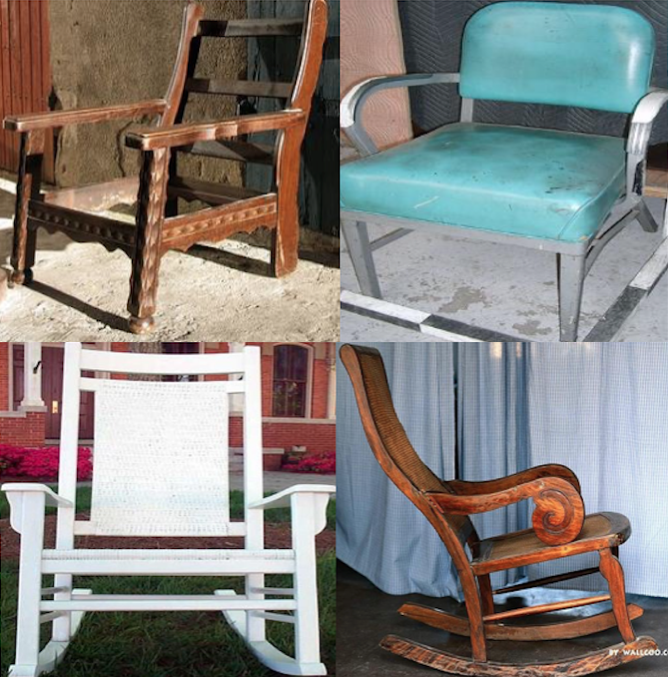}
		\caption{Clean background \& cluttered background}
		\label{fig:clean_clutter}
	\end{figure}

	Rendered image test set, as we mentioned in Sec. \ref{trainingdata}, consists of $1057 \times 16$ rendered images, which also comes from ShapeNet. Clean background and cluttered background real image test sets are collected from ImageNet $[\ref{bib:imagenet}]$, containing 1309 and 1000 images respectively, both with manually labeled ground truth of viewpoint. Some sample images are shown in Figure \ref{fig:clean_clutter}. Obviously, these three datasets are increasingly noisy and thus difficult to tackle.\\
	
	For image preprocessing and feature extraction on the test sets, we used the same scheme as the training set. That is, convert each image into a 20736-dimensional HoG feature.
	
\section{Model}\label{model}
Rather than using global image feature as the input of classification, our pose estimation model is patch-based. By dividing image into patches and training a classifier for each patch, our model can be more robust to occlusion and background noise. Also, this approach reduced the feature dimension for each classifier, thus reduced the sample complexity. Actually, we did try global features, while the classification accuracy is 30\% lower than patch based method on clean background test set, shown in Table \ref{tab:result}. The mathematical representation of our patch based model is as follows.\\

Define $F_i$ as the HoG feature of patch $i$, $I=(F_1,\cdots,F_{N^2})$ to be the HoG feature of the whole image, $\mathcal{V} = \{1,\cdots,V\}$ to be the discretized pose space. For each patch, we build a classifier, which learns from training data, and gives a prediction of the conditional probability $P(v|F_i)$. To respresent $P(v|I)$ in $P(v|F_i)$, $i=1,\cdots,N^2$, we assume $P(v|I) \propto \prod\limits_{i = 1}^{{N^2}} {P(v|{F_i})}$. So, we can calculate $P(v|I)$ and the according $\bar{v}$ using the following formula:\\
\[P(v|I) = \frac{\prod\limits_{i = 1}^{{N^2}} {P(v|{F_i})}}{\sum\limits_{v=1}^{V}\prod\limits_{i = 1}^{{N^2}} {P(v|{F_i})}}\]
\[\bar{v} = \arg\max_{v} P(v|I)\]

In sum, our model takes $F_i$, $i=1,\cdots,N^2$ as input, and outputs $P(v|I)$ and $\bar{v}$.

\section{Methods}\label{methods}
	\subsection{Learning Algorithms}
		\subsubsection{Random Forest}
		In this project, we choose random forest [\ref{bib:rf}] as a primary classification algorithm based on its following advantages:
		\begin{itemize}
			\item Suitable for multiclass classification.
			\item Non-parametric, easy to tune.
			\item Fast, easy to parallel.
			\item Robust, due to randomized processing.
		\end{itemize}
		
		During classification, 36 random forest classifiers are trained for 36 patches. As a trade off between spatio-temporal complexity and performance, we set the forest size to be 100 trees. We also tuned the maximum depth of trees using cross-validation, where the optimal depth is 20. As a result, each random forest outputs a probability vector $P(v|F_i)$. After Laplace smoothing, we calculated $P(v|I)$, estimated the pose to be $\bar{v} = \arg\max\limits_{v}{P(v|I)}$. 
		
	\subsection{Optimization}
	Constructing training dataset from rendered images has many advantages, but there are also drawbacks. As I mentioned in Section \ref{intro}, the prior probability of pose in real images can be highly different from that in rendered images. As we know, pose distribution in the training set is uniform, however, in real images, there are far more front view chairs than back view. Fortunately, this difference can be analyzed and modeled as follows.
	
	\subsubsection{Probability Calibration}
	In classification step, each classifier $C_i$ will output a probability vector $\tilde{P}(v|F_i)$. Using Bayesian formula, we have:
	\[\tilde{P}(v|F_i) = \frac{\tilde{P}(v)\tilde{P}(F_i|v)}{\tilde{P}(F_i)}\]
	
	Although we may not learn $\tilde{P}(v)$, $\tilde{P}(F|v)$ and $\tilde{P}(F)$ explicitly when training, we can use them to indicate the statistical property of training data. Whereas, the real $P(v|F_i)$, which satisfies the following formula, could be different from $\tilde{P}(v|F_i)$. Here, $P(v)$, $P(F_i|v)$ and $P(F_i)$ are statistical properties of the test set.
	\[P(v|F_i) = \frac{P(v)P(F_i|v)}{P(F_i)}\]
	
	Assume the training data and the test data have at least some similarity. Specifically speaking, assume $P(F_i|v) = \tilde{P}(F_i|v)$, $P(F_i) = \tilde{P}(F_i)$, then we have:
	\[P(v|F_i) = \tilde{P}(v|F_i) \frac{P(v)}{\tilde{P}(v)} \propto \tilde{P}(v|F_i) P(v)\]
	
	To recover $P(v|F_i)$, we just need to achieve a good estimation of $P(v)$. One possible method might be randomly choosing some samples from the test set, and manually label the ground truth of viewpoint, regard the ground truth pose distribution of samples as an estimation of overall $P(v)$. However, we still need to do some ``labor work'' --- labeling.\\
	
	Noticing the above formula can also be written as:
	\[\frac{P(v|F_i)}{P(v)} = \frac{\tilde{P}(v|F_i)}{\tilde{P}(v)} \mbox{; }\tilde{P}(v) = \frac{1}{V} \mbox{, }\forall v \in \mathcal{V}\]
	we came up with another idea to automatically improve the classification result. For $\tilde{P}(v|F_i)$, we have:
	\[P(v)>\frac{1}{V} \Rightarrow \tilde{P}(v|F_i) < P(v|F_i)\]
	\[P(v)<\frac{1}{V} \Rightarrow \tilde{P}(v|F_i) > P(v|F_i)\]
	That means, when testing, frequently appeared poses are underestimated, while uncommon poses are overestimated. Here, we propose an iterative method to counterbalance this effect. Basically, we will use $\tilde{P}(v|F_i)$ to generate an estimation $\tilde{P}(v)$ of the prior distribution; assume $P(v)$ and $\tilde{P}(v)$ have similar common views and uncommon views (in other words, $P(v)$ and $\tilde{P}(v)$ have the same trend); smooth $\tilde{P}(v)$ to keep the trend while reduce fluctuation range; multiply the original $\tilde{P}(v|F_i)$ by smoothed $\tilde{P}(v)$; and iteratively repeat the above steps. Finally, due to the damping effect in combination step, $\tilde{P}(v)$ will converge, and $\tilde{P}(v|F_i)$ gets closer to $P(v|F_i)$. Formulation of this iterative algorithm is as follows:\\

	\begin{enumerate}
		\item Calculate $\tilde{P}(v|I^{(j)})$, $j=1,\cdots,m$.
		\[\tilde{P}(v|I^{(j)}) = \frac{\prod\limits_{i = 1}^{{N^2}} {\tilde{P}(v|{F_i^{(j)}})}}{\sum\limits_{v=1}^{V}\prod\limits_{i = 1}^{{N^2}} {\tilde{P}(v|{F_i^{(j)}})}}\]
		\item Accumulate $\tilde{P}(v|I^{(j)})$ on all test samples to calculate $\tilde{P}(v)$.
		\[\tilde{P}(v) = \frac{1}{m}\sum_{j=1}^{m}{\tilde{P}(v|I^{(j)})}\]
		\item Smooth $\tilde{P}(v)$ by factor $\alpha$.
		\[\tilde{P}_{\rm s}(v)=\frac{\tilde{P}(v)+\alpha}{1+16\alpha}\]
		\item Estimate $P(v|F_i)$ by letting:
		\[\bar{P}(v|F_i) = \tilde{P}(v|F_i) \tilde{P}_{\rm s}(v)\]
		\item Use $\bar{P}(v|F_i)$ to re-calculate $\tilde{P}(v|I^{(j)})$ in step 1, while remain $\tilde{P}(v|F_i)$ in step 4 unchanged, repeat the above steps.
	\end{enumerate}

	\subsubsection{Parameter Automatic Selection}	
\begin{figure}[h]
		\centering
		\includegraphics[width=0.40\textwidth]{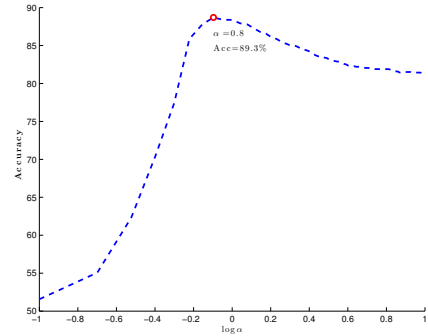}
		\caption{Classification accuracy change w.r.t. $\alpha$}
		\label{fig:acc_logalpha}
	\end{figure}

\begin{figure}[h]
		\centering
		\includegraphics[width=0.40\textwidth]{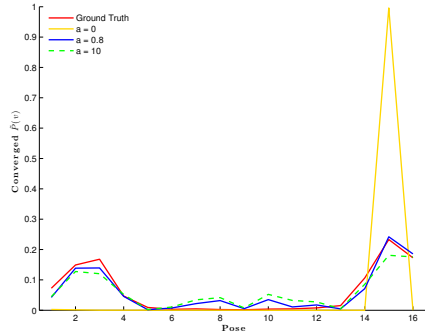}
		\caption{Stable distribution $\tilde{P}(v)$ w.r.t. $\alpha$}
		\label{fig:distribution_alpha}
	\end{figure}

	After several iterations, the algorithm will converge, and we'll get a final estimation $\bar{P}(v|F_i)$ of $P(v|F_i)$. However, different $\alpha$ will lead to far different converging results, as shown in Figure \ref{fig:acc_logalpha}. From experiment results in Figure \ref{fig:distribution_alpha} we observed that if $\alpha$ is too small, viewpoint with the highest initial probability $\tilde{P}(v)$ will soon beat other viewpoints, and $\tilde{P}(v)$ converges to a totally biased distribution. While, if $\alpha$ is too large, smoothing effect is too strong to influence $\bar{P}(v|F_i)$, resulting in $\bar{P}(v|F_i) = \tilde{P}(v|F_i)$. However, there exists an intermediate value of $\alpha$ to maximize the classification accuracy and lead to an optimal estimation $\bar{P}(v|F_i)$. In Figure \ref{fig:acc_logalpha} and \ref{fig:distribution_alpha}, $\alpha_{\rm opt}$ is 0.8.\\
	
	
\begin{figure}[h]
		\centering
		\includegraphics[width=0.50\textwidth]{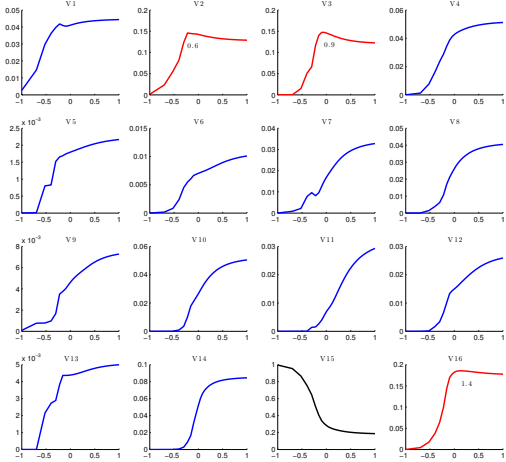}
		\caption{$\tilde{P}(v_j)$ curve with respect to $\alpha$}
		\label{fig:p_alpha}
	\end{figure}

\begin{figure}[h]
		\centering
		\includegraphics[width=0.40\textwidth]{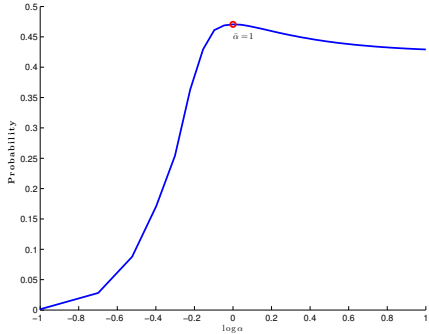}
		\caption{Estimated $\alpha$}
		\label{fig:est_alpha}
	\end{figure}

	To solve the optimal $\alpha$, we conducted deep analysis to the relationship between stable $\tilde{P}(v)$ and $\alpha$. We found three patterns of relationship between $\tilde{P}(v_j)$ and $\alpha$, shown in Figure \ref{fig:p_alpha}. For some viewpoints, $\tilde{P}(v)$ is almost monotonically increasing with respect to $\alpha$, such as blue curves, some are monotonically decreasing, such as the black curve, while others will decrease after first increase, such as the red curves. Recall the distribution change with respect to $\alpha$ in Figure \ref{fig:distribution_alpha}, we found $\tilde{P}(v)$ will first approximate $P(v)$ then be smoothed. Therefore, patterns with turning points are good reflection of this trend. Sum on those components, we get Figure \ref{fig:est_alpha}, and take the turning point of the curve as our estimated $\bar{\alpha}$. Here $\bar{\alpha}$ is 1, very close the optimal value $\alpha_{\rm opt} = 0.8$.
	
	
\section{Results}\label{results}
	\subsection{Classification Performance}
	\begin{table}[h]
		\centering
		\begin{tabular}{c|c|c|c}
			\hline
			\hline
			& Render & Clean & Cluttered\\
			\hline
			RF(\%) & 96.16 & 80.67 & 76.80\\
			\hline
			RF$_{\rm opt}$(\%) & --- & \textbf{88.90} & \textbf{78.70}\\
			\hline
			RF$_{\rm GT}$(\%) & --- & 91.29 & 81.00\\
			\hline
			Global(\%) & \textbf{97.03} & 52.64 & 10.90\\
			\hline
			\hline
		\end{tabular}
		\caption{Classification accuracy on three test sets}
		\label{tab:result}
	\end{table}
	In Table \ref{tab:result}, our patch based random forest classification algorithm (denoted as RF) shows a promising classification results on all three test sets. Under our scheme, random forest achieves $80\%$ accuracy on clean background real image test set, and $77\%$ on cluttered background test set. After calibrating the conditional probability $\tilde{P}(v|F_i)$ using automatically selected $\alpha$ (denoted as RF$_{\rm opt}$), performance on clean test set is boosted by $8\%$, as well $2\%$ on cluttered set. The relatively low improvement on cluttered test set may result from our assumption of $\tilde{P}(F_i|v)=P(F_i|v)$ and $\tilde{P}(F_i)=P(F_i)$ are too strong for cluttered images.\\
	
	Row ``RF$_{\rm GT}$'' shows the result of calibrating $\tilde{P}(v|F_i)$ using ground truth $P(v)$. Compared to our optimization approach, the accuracy is only 2\% higher, indicating the effectiveness of our method.\\
	
\begin{figure}[h]
		\centering
		\includegraphics[width=0.5\textwidth]{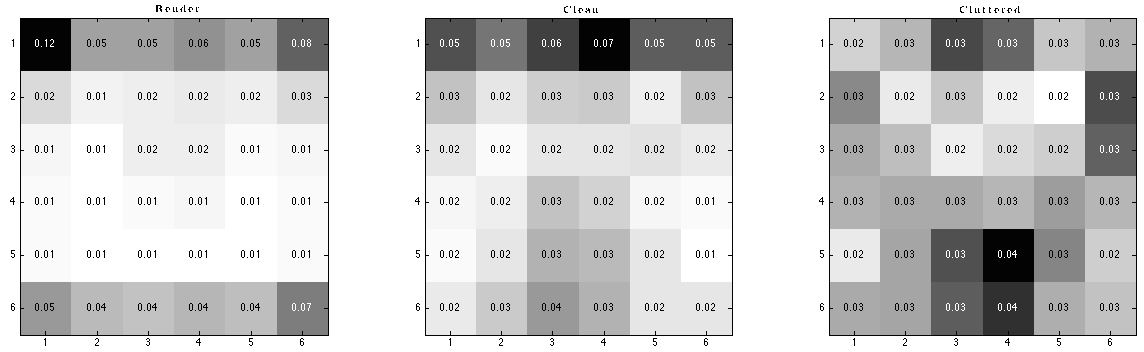}
		\caption{Patch importance on rendered, clean, and cluttered test sets. Learned by training a global random forest classifier on three datasets.}
		\label{fig:feat_imp}
	\end{figure}

	Besides, the ``Global'' row shows a terrible classification performance on global image features. Although it achieves best result on rendered images, performance drops significantly when testing on real images. One possible explanation might be that global classifier overfittingly learned the importance of patches from training set, while patch importance on real images is different. Figure \ref{fig:feat_imp} verified this hypothesis. In contrast, patch based method gives equal importance to all patches, hence reduced overfitting.\\
		
\begin{figure}[h]
		\centering
		\includegraphics[width=0.5\textwidth]{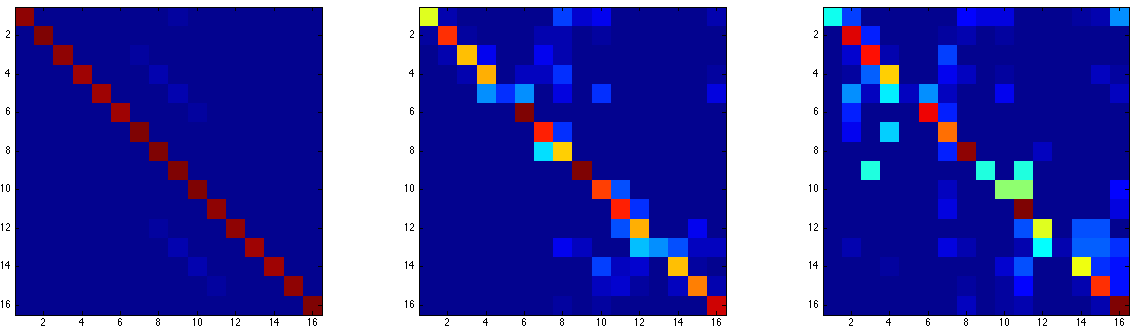}
		\caption{Confusion matrix on rendered, clean, and cluttered test sets}
		\label{fig:confusion_matrix}
	\end{figure}

	Figure \ref{fig:confusion_matrix} shows the confusion matrix on three test sets respectively. From left to right, as test difficulty increases, confusion matrix becomes increasingly scattered. On rendered image test set, an interesting phenomenon is that some poses are often misclassified to poses with $90^\circ$ difference with them, one possible explanation is that the shape of some chairs are like a square. Also, front view and backview are often misclassified, because they have similar appearance in feature space.
	
\section{Conclusion}\label{conclusion}
In this paper, we proposed a novel pose estimation approach --- learn from 3D models. We explained our model in Bayesian framework, and raised a new optimization method to transmit information from 2D images to 3D models. The promising experiment results verified the effectiveness of our scheme.
	
\section{Future Work}\label{discussion}	
	We have several ideas for the future work, described as follows:
	\begin{itemize}
		\item Take into consideration the foreground and background information in the image, fully utilize the information in rendered images.
		\item Further model the difference between three datasets, revise our inaccurate assumption.
		\item Learn the discriminativeness of patches, give different weight for different patches.
		\item Generalize our algorithm to occluded images, or different categories, see what will happen.
	\end{itemize}

\bibliographystyle{abbrv}

\end{document}